\documentclass[10pt,letterpaper,twocolumn]{article}

\usepackage[margin=0.75in]{geometry}
\usepackage{amsmath}
\usepackage{amssymb}
\usepackage{booktabs}
\usepackage{caption}
\usepackage{graphicx}
\usepackage{makecell}
\usepackage{multirow}
\usepackage{pifont}
\usepackage{stfloats}
\usepackage{xcolor}
\usepackage[hidelinks]{hyperref}

\newcommand{\cmark}{\textcolor{green!45!black}{\ding{51}}}
\newcommand{\xmark}{\textcolor{red!80!black}{\ding{55}}}

\newcommand{\compacttable}{%
    \footnotesize
    \setlength{\tabcolsep}{3pt}%
    \renewcommand{\arraystretch}{1.0}%
    \setlength{\aboverulesep}{0.3ex}%
    \setlength{\belowrulesep}{0.3ex}%
}

\title{What the Elevation Map Cannot See:
Semantic-Aware Locomotion and Execution-Aware Navigation for Humanoid Robot}

\author{%
    \parbox{0.97\textwidth}{%
        \centering
        Shunyu Yao$^{1}$, Songyang Liu$^{1}$, Dinghao Chen$^{2}$,
        Yuanyuan Lei$^{3}$, and Shuai Li$^{1}$\\[0.75em]
        {\small
        $^{1}$Engineering School of Sustainable Infrastructure \& Environment,\\
        University of Florida, Gainesville, FL 32611, USA\\
        $^{2}$Independent researcher\\
        $^{3}$Department of Computer \& Information Science \& Engineering,\\
        University of Florida, Gainesville, FL 32611, USA}\\[0.5em]
        {\footnotesize
        \texttt{shunyu.yao@ufl.edu}, \texttt{liusongyang@ufl.edu},
        \texttt{shuai.li@essie.ufl.edu}\\
        \texttt{dinghaochen@ucsb.edu}, \texttt{yuanyuan.lei@ufl.edu}}%
    }%
}
\date{}

\begin{document}

\twocolumn[{%
    \maketitle
    \begin{center}
        \begin{minipage}{0.97\textwidth}
            \centering
            \includegraphics[width=0.93\linewidth]{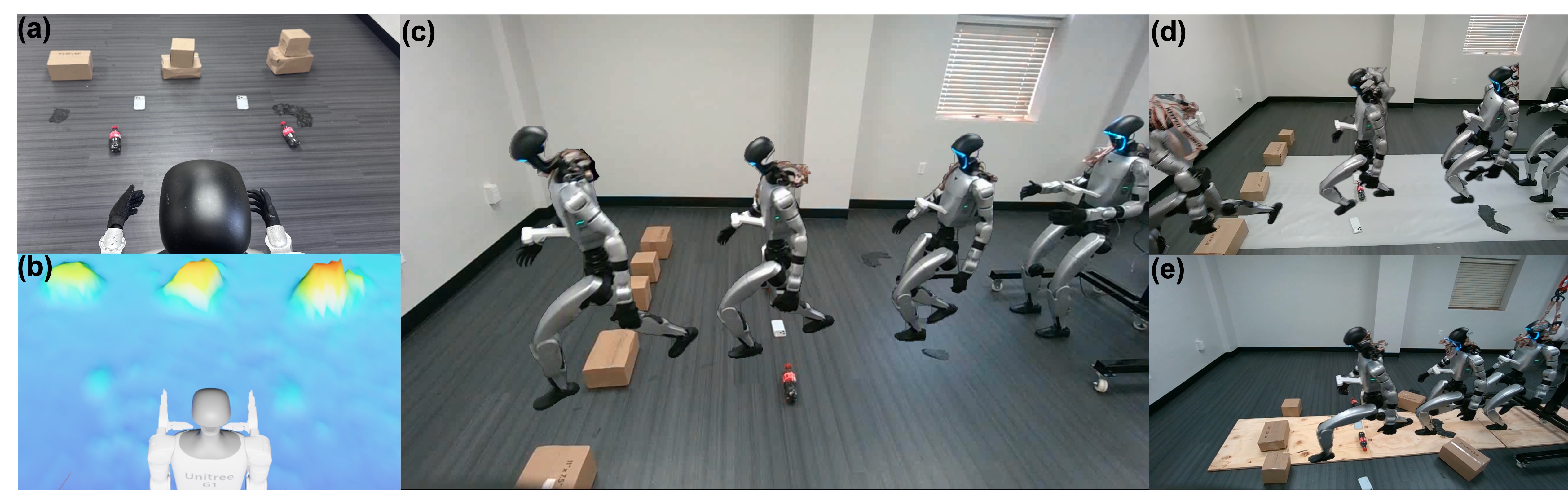}
            \captionsetup{hypcap=false}
            \captionof{figure}{Hazard visibility and real-world locomotion experiments.
            (a) All obstacles and hazards are visible in the RGB image.
            (b) In the map reconstructed using FAST-LIO2~\cite{xu2022fast}, the box is clearly visible (elevation-visible),
            whereas the bottle is difficult to distinguish from noise on the ground
            (elevation-ambiguous); the phone and water spill are not discernible
            (elevation-invisible).
            (c)--(e) Results of real-robot experiments with our locomotion policy on three different terrain types.}
            \label{fig:teaser}
        \end{minipage}
    \end{center}
    \bigskip
}]

\begin{abstract}

Navigation for humanoid robots is critical, yet large-scale evaluation on physical hardware is often impractical due to cost and safety concerns, making simulation benchmarks essential. Existing VLN benchmarks achieve physically executable navigation, but still assume (1) all hazards are observable from elevation maps; (2) realized motions closely match desired motions. In real environments, however, fallen bottles may be ambiguous in elevation maps, while phones and water spills may be difficult to differentiate; hazard avoidance by the locomotion policy can cause the robot's actual trajectory to deviate from the path intended by the VLN policy. Such command-execution mismatch can accumulate and lead the robot toward unintended locations. To expose these failure modes, we introduce a benchmark that models both elevation-subtle hazards and execution deviations, together with a closed-loop VLN + locomotion control framework that continuously realigns high-level navigation with the robot's actual state. We evaluate navigation in simulation and further validate the locomotion policy on a physical Unitree G1 humanoid robot. Results show that semantic input reduces contact with hazards poorly represented in elevation maps, while anti-deviation improves navigation success. These findings highlight the need to evaluate humanoid navigation jointly in terms of route completion and hazard avoidance. Project page: \url{https://jas000n.github.io/elevation-map-cannot-see/}.

\end{abstract}

\section{Introduction}

Humanoid robots are increasingly expected to assist people in everyday household environments, where navigation requires moving through rooms and corridors while avoiding furniture and floor-level hazards. Small objects and water spills in these environments may be poorly represented in depth-derived elevation maps, posing challenges beyond geometric obstacle avoidance. Although humanoids offer mobility advantages over wheeled platforms, large-scale real-world evaluation is costly, time-consuming, and potentially unsafe. Simulation benchmarks therefore enable repeatable, scalable evaluation before physical deployment.

Vision-and-Language Navigation (VLN) allows robots to navigate according to natural-language instructions rather than predefined metric goals. Early VLN benchmarks moved agents between predefined viewpoints, abstracting away locomotion and obstacle interaction \cite{anderson2018vision}. Continuous-environment VLN replaced this graph-based formulation with local actions in continuous 3D environments \cite{krantz2020beyond}. More recently, physics-based benchmarks such as VLN-PE~\cite{wang2025rethinking}, VLNVerse~\cite{lin2025vlnverse}, and VLN-CE-Isaac~\cite{cheng2025navila} further introduce robot kinematics, collision dynamics, and continuous motion, advancing VLN toward physically executable navigation.

Nevertheless, physics-based VLN evaluation retains two implicit assumptions. First, obstacle observability assumes that locomotion hazards are detectable from depth or elevation maps. Yet wet floors and small objects such as phones or cables may have weak or absent geometric signatures, so geometric traversability does not imply safety. Second, command-execution consistency assumes that realized motion matches high-level commands. Safe humanoid locomotion may instead require sidesteps, shorter steps, or turns. Repeated avoidance can accumulate pose and heading errors, shift observations away from those expected by the VLN policy, and undermine navigation.

These limitations are coupled: avoiding a semantically detected hazard can alter the trajectory assumed by the VLN policy. Existing benchmarks emphasize geometric traversability and collisions rather than this perception-execution feedback loop. Strong simulation performance may therefore conceal missed non-geometric hazards or lost task progress when subsequent decisions ignore avoidance-induced deviations.

We introduce a humanoid VLN benchmark that exposes these failure modes through hazards with different levels of geometric observability, including those weakly represented or absent in elevation maps. It distinguishes commanded from realized motion, enabling controlled evaluation of execution deviations caused by safety-aware low-level behavior. This formulation measures whether an agent reaches its language-specified destination while remaining safe and recovering its navigation state after local deviations.

We develop a hierarchical framework coupling a high-level VLN module with a semantic-aware whole-body locomotion policy. The VLN module generates relative target poses from language instructions and visual observations, while the locomotion policy tracks the derived velocity commands with local terrain and hazard awareness. Closed-loop feedback of the robot's realized state corrects avoidance-induced deviations. We evaluate navigation in simulation and validate locomotion on a physical Unitree G1 with hazards poorly represented by geometry alone.

\begin{figure*}[t]
    \centering
    \includegraphics[width=0.7\linewidth]{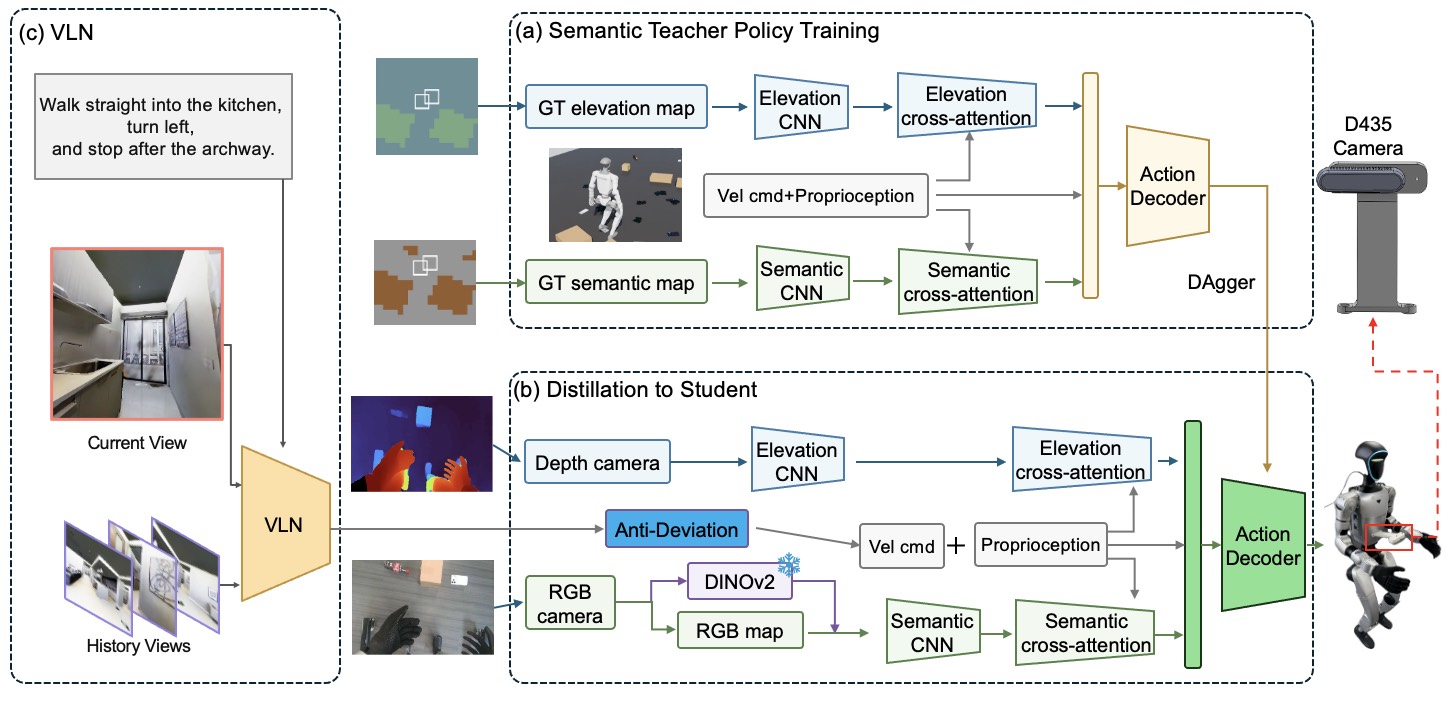}
    \caption{Method overview. (a) A semantic-aware teacher learns locomotion from ground-truth elevation and semantic maps. (b) DAgger distills the teacher into an RGB-D student with elevation and semantic branches. (c) We use the high-level VLN module from NaVILA~\cite{cheng2025navila} to generate relative target poses, which the anti-deviation module helps track using localization feedback.}
    \label{fig:methodology_overview}
\end{figure*}

\begin{table}[t]
    \centering
    \small
    \setlength{\tabcolsep}{5pt}
    \caption{Comparison of Navigation Benchmarks}
    \label{tab:benchmark_comparison}
    \resizebox{\columnwidth}{!}{%
    \begin{tabular}{llcccccc}
        \toprule
        Benchmark & Simulator & VLN & Physics & Humanoid & \makecell{Geometric obstacles \\on path} & \makecell{Semantic hazards \\on path} & \makecell{Dynamics variations\\\& external disturbances} \\
        \midrule
        VLN-CE~\cite{krantz2020beyond} & Habitat & \cmark & \xmark & \xmark & \xmark & \xmark & \xmark \\
        HumanoidBench~\cite{sferrazza2024humanoidbench} & MuJoCo & \xmark & \cmark & \cmark & \xmark & \xmark & \xmark \\
        VLN-PE~\cite{wang2025rethinking} & Isaac Sim & \cmark & \cmark & \cmark & \xmark & \xmark & \xmark \\
        VLN-CE-Isaac~\cite{cheng2025navila} & Isaac Sim & \cmark & \cmark & \cmark & \xmark & \xmark & \xmark \\
        \midrule
        \textbf{Ours} & Isaac Sim & \cmark & \cmark & \cmark & \cmark & \cmark & \cmark \\
        \bottomrule
    \end{tabular}%
    }
\end{table}

Our contributions are two-fold:
\begin{itemize}
        \item Semantic-aware locomotion + anti-deviation VLN: The semantic branch provides a "steppability" prior to fill the blind spots of the elevation map, while anti-deviation correction compensates local
deviations before they accumulate into navigation drift or
render subsequent commands unsafe.
        \item Hazard taxonomy and benchmark: Incorporate "geometrically invisible hazards" and "physical disturbances" into the VLN evaluation benchmark, complemented by new metrics such as the Hazard Contact Rate. Evaluation: Navigation is evaluated in simulation, while locomotion is evaluated in both simulated and real-world environments.
\end{itemize}

\section{Related Works}

\subsection{Humanoid Navigation Benchmark}

VLN datasets such as RxR \cite{ku2020rxr} and REVERIE \cite{qi2020reverie} extend language and object grounding without modeling humanoid execution. HumanoidBench \cite{sferrazza2024humanoidbench} evaluates physical control but not language-guided navigation. Recent physics-based VLN benchmarks incorporate robot embodiment and locomotion \cite{wang2025rethinking,cheng2025navila,lin2025vlnverse,humanoidvln}, complementing these language- and control-focused evaluations.

These benchmarks advance physical evaluation but cover a more limited range of perturbations and obstacles. We introduce richer perturbation and obstacle settings to jointly diagnose semantic hazards and safety-induced execution deviations. We target hazards poorly represented in elevation maps, such as wet surfaces and small objects, and evaluate safety and navigation recovery when low-level avoidance alters commanded motion.

\subsection{Humanoid Locomotion and Whole-Body Control}

Perpetual Humanoid Control (PHC) \cite{luo2023perpetual} provides motion imitation and fall recovery for simulated avatars. ExBody \cite{cheng2024expressive} combines upper-body imitation with velocity-tracking locomotion on hardware, while OmniH2O \cite{he2024omnih2o} learns deployable whole-body control from retargeted human motion and privileged teachers. HOVER \cite{he2025hover} unifies velocity, pose, and upper-body control interfaces; ExBody2 \cite{ji2026exbody2} and TWIST \cite{ze2025twist} emphasize motion tracking and coordinated locomotion--manipulation. Their primary focus is reference tracking rather than semantic navigation hazards. Terrain-aware methods incorporate environmental geometry. Humanoid Parkour \cite{zhuang2025parkour} uses vision-based reinforcement learning for obstacle traversal. Attention-Based Map Encoding (AME) \cite{he2025attention} attends to elevation maps for foothold selection, and AME-2 \cite{zhang2026ame2} adds uncertainty-aware mapping. Hiking in the Wild \cite{zhu2026hiking} instead maps depth and proprioception directly to humanoid joint actions. Geometry alone, however, cannot distinguish similarly shaped surfaces with different semantic risks.

Related quadruped systems address complementary aspects: ViNL \cite{kareer2023vinl} couples visual navigation with locomotion that steps over household clutter, while Yang et al.\ \cite{yang2023semantics} learn terrain-semantic speed and gait selection from human demonstrations. Our focus is humanoid foot-level hazard avoidance using aligned geometric and semantic maps, coupled with realized-motion feedback for correcting deviations during language-guided navigation.

\section{Methodology}

\begin{figure}[tbp]
    \centering
    \includegraphics[width=0.7\linewidth]{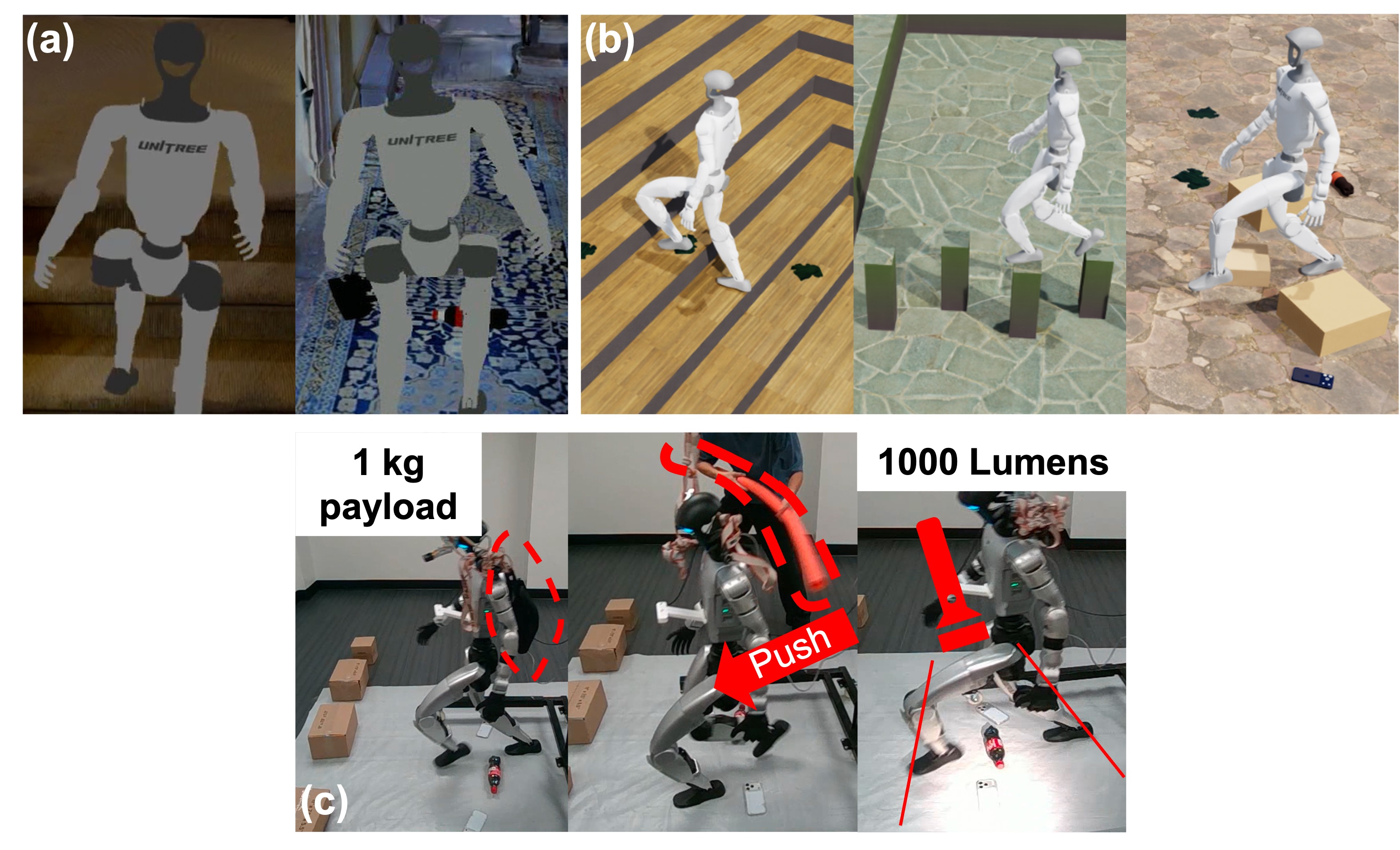}
    \caption{Testing environments. (a) Navigation in simulation; (b) Locomotion in simulation; (c) Real-world with external payload/push and lighting conditions.}
    \label{fig:testing_environments}
\end{figure}

\subsection{Problem Formulation}

We formulate humanoid navigation among semantically identifiable hazards
with ambiguous or absent elevation signatures as a hierarchical task.
At decision step $t$, the high-level VLN policy maps a language instruction
$I$, the current RGB image $o_t^{\mathrm{nav}}$, and the previous $L$ images
$\mathcal{H}_t=\{o_{t-L}^{\mathrm{nav}},\ldots,o_{t-1}^{\mathrm{nav}}\}$
to a relative target pose in the robot frame:
\begin{equation}
    w_t = \left(\Delta x_t,\Delta y_t,\Delta\theta_t\right)
    = \pi_{\mathrm{vln}}\!\left(I,o_t^{\mathrm{nav}},\mathcal{H}_t\right)
    \in\mathbb{R}^{3}.
    \label{eq:nav_policy}
\end{equation}
Here, $\Delta x_t$ and $\Delta y_t$ are planar displacements and
$\Delta\theta_t$ is a heading change. Composing $w_t$ with the robot's planar pose $p_t$ at the decision
time gives the intended target pose $p_t^{\mathrm{tar}}=p_t\oplus w_t$,
i.e., the pose the robot would reach if $w_t$ were executed exactly.
The locomotion policy does not consume $w_t$ directly; it tracks a planar
velocity command $c_t=(v_{x,t},v_{y,t},\psi_t)$ derived from $w_t$ as described
in Sec.~\ref{sec:anti_deviation}, where $v_{x,t},v_{y,t}$ are linear velocities
and $\psi_t$ is the yaw rate. The low-level policy directly consumes
waist-mounted RGB-D images $(o^{\mathrm{rgb}},o^{\mathrm{depth}})$ and the
proprioceptive input
\begin{equation}
    r = \left[\mathbf{v}^{\mathrm{base}},\,g,\,c_t,\,q,\,\dot q,\,a^{\mathrm{prev}}\right]
    \in \mathbb{R}^{96}.
    \label{eq:loco_observation}
\end{equation}
Here, $\mathbf{v}^{\mathrm{base}},g\in\mathbb{R}^{3}$ denote planar base
velocity and projected gravity, and $q,\dot q,a^{\mathrm{prev}}\in\mathbb{R}^{29}$
are joint positions, joint velocities, and the previous action.
The policy outputs joint-position offsets from a default pose $q^{\mathrm{default}}$:
\begin{equation}
\begin{aligned}
    a &= \pi_{\mathrm{loco}}\!\left(
    o^{\mathrm{rgb}},o^{\mathrm{depth}},r\right)
    \in\mathbb{R}^{29},
    q^{\mathrm{target}} &= q^{\mathrm{default}} + a.
\end{aligned}
    \label{eq:loco_policy}
\end{equation}

The objective is to reach the language-specified goal without hazard
contact or falls, while keeping the realized pose close to the
VLN-commanded target poses so that local hazard avoidance does not
accumulate into navigation drift.

\begin{figure*}[t]
    \centering
    \includegraphics[width=0.65\linewidth]{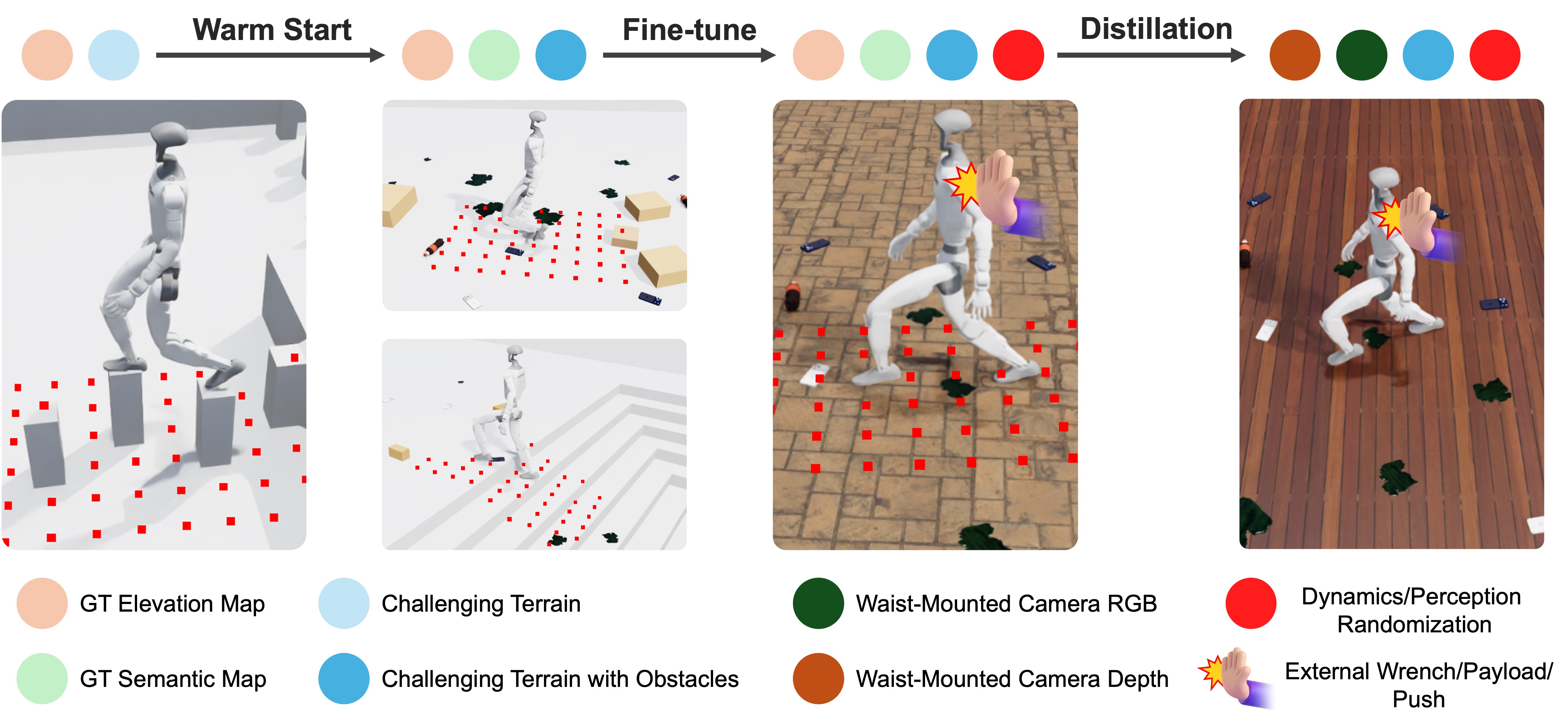}
    \caption{Training curriculum for semantic-aware locomotion. An elevation-based policy provides a warm start for learning hazard avoidance using ground-truth elevation and semantic maps on challenging terrains with obstacles. The teacher is then fine-tuned under domain randomization and external disturbances to improve robustness. Finally, teacher-student distillation transfers this capability to a deployable policy using waist-mounted RGB-D observations. Colored markers indicate the observations and training conditions used at each stage.}
    \label{fig:stage}
\end{figure*}

\subsection{Semantic Branch for Privileged Locomotion}
\label{sec:semantic_branch}
Fig.~\ref{fig:stage} shows our locomotion training curriculum. We first learn how to use semantic hazard information for locomotion
under ground-truth perception, then transfer this capability to a waist
camera-based student.

\subsubsection{Augmenting the geometric locomotion policy}
We start from a pretrained AME policy~\cite{he2025attention} that can
already walk and traverse challenging terrain. We retain its network architecture and weights, and add
an independent semantic branch.
The geometric branch describes terrain shape, whereas the semantic
branch provides a steppability representation that distinguishes safe
support from hazardous regions. The teacher generates actions
using ground-truth elevation and semantic maps together with proprioceptive input.

In the first stage, we augment the challenging terrain with hazards that are visible, ambiguous and invisible in elevation map.
We train the augmented policy using PPO with ground-truth maps and two
additional reward terms: a step-over clearance reward and a premature-descent
penalty (Sec.~\ref{sec:step_over_rewards}). This stage builds semantic-aware
hazard avoidance on top of the pretrained terrain-traversal capability.

\subsubsection{Training a robust semantic teacher}
We continue PPO training with curriculum-based
randomization of dynamics and payload, proprioceptive observation noise,
pushes, and external forces, as detailed in
Sec.~\ref{sec:experiment_setup}. The policy retains ground-truth elevation
and semantic maps while learning to handle these perturbations, yielding a
robust semantic teacher.

\subsubsection{Distillation for Deployable RGB-D Locomotion}
\label{sec:rgbd_distillation}
\begin{figure}[tbp]
    \centering
    \includegraphics[width=0.8\linewidth]{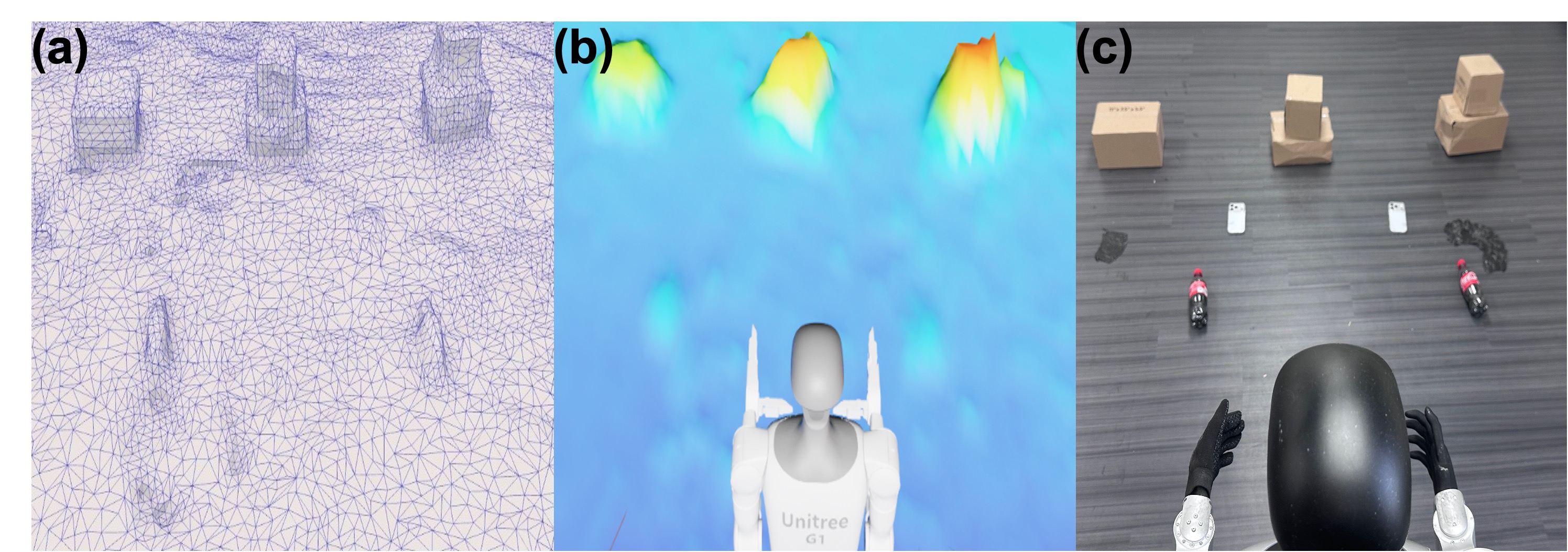}
    \caption{(a) Terrain mesh from detailed offline handheld scanning, (b) reconstructed elevation map, and (c) RGB observation.}
    \label{fig:perception_gap}
\end{figure}

The teacher in Section~\ref{sec:semantic_branch} already uses semantics
to guide locomotion, but depends on ground-truth maps unavailable on
hardware. As shown in Fig.~\ref{fig:perception_gap}, small hazards may be weakly represented in reconstructed elevation
maps, while the safety relevance of geometrically flat hazards may not be
apparent from height information alone. RGB observations provide complementary
appearance cues for identifying such hazards. We therefore transfer the teacher's
control behavior to a student using onboard RGB-D and proprioception,
while retaining the action interface.

\paragraph{DINO-augmented semantic perception}
We construct a new student from the robust teacher. Its geometric branch receives camera depth, while its semantic branch receives RGB and
additional visual features extracted by a
frozen DINOv2-S backbone~\cite{oquab2023dinov2}. A trainable projection converts the DINO
features into additional channels concatenated with the semantic CNN
input. At deployment, we use
Fast-FoundationStereo (FFS)~\cite{wen2026fast}
to obtain cleaner depth observations from stereo image pairs for the geometric branch.

\begin{figure}[tbp]
    \centering
    \includegraphics[width=0.6\linewidth]{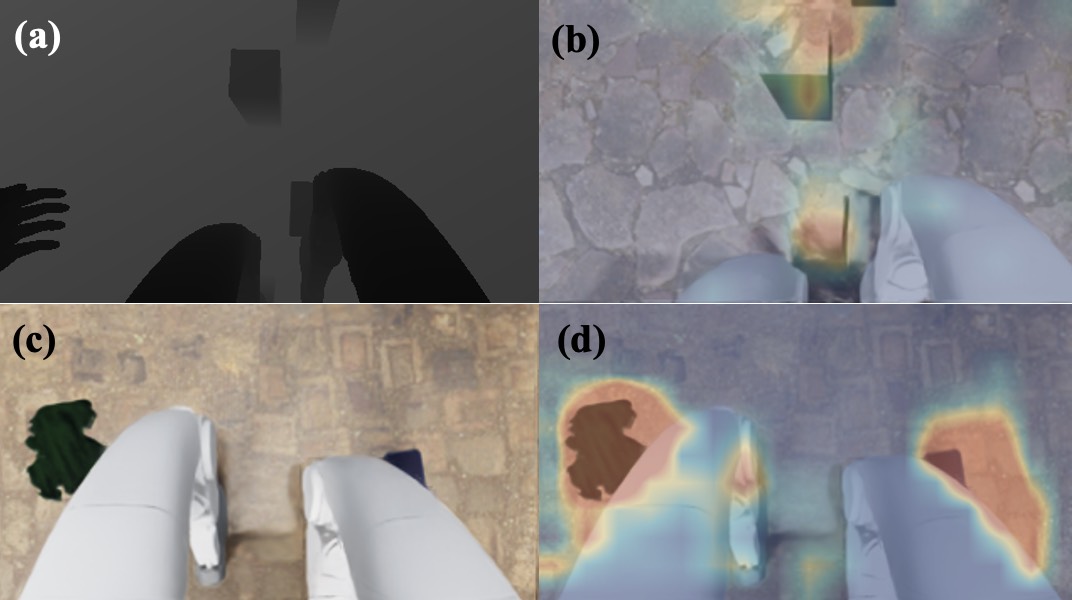}
    \caption{Selected Grad-CAM visualizations conditioned on right-leg actions.
    (a) Depth observation during stake traversal.
    (b) Corresponding Grad-CAM visualization.
    (c) RGB observation while stepping over semantic obstacles.
    (d) Corresponding Grad-CAM visualization.}
    \label{fig:grad_cam}
\end{figure}

Fig.~\ref{fig:grad_cam} presents two Grad-CAM examples conditioned on
right-leg actions. The highlighted regions lie on walkable supporting
surfaces during stake traversal and on semantic hazards that require
avoidance during obstacle crossing. In these examples, the salient
responses focus on the surrounding terrain and hazards rather than
the robot's own legs.

\paragraph{Action distillation with dataset aggregation}
We distill the frozen gt map-based teacher into the RGB-D student using
DAgger, minimizing action-mean squared error at matched states:
\begin{equation}
\begin{aligned}
    \mathcal{L}_{\mathrm{distill}}
    &= \mathbb{E}_{s\sim\mathcal{D}_{\mathrm{mix}}}
       \left[\frac{1}{29}\left\|
       \mu_S(o_s^{\mathrm{rgb}},o_s^{\mathrm{depth}},r_s)
       \right.\right.\\
    &\hspace{5em}\left.\left.
       -\mu_T(E_s,S_s,r_s)\right\|_2^2\right],
\end{aligned}
\label{eq:action_distillation}
\end{equation}
where $\mathcal{D}_{\mathrm{mix}}$ contains states from teacher--student
rollouts. The teacher controls the first 200 iterations, then control
transfers gradually to the student over 800 iterations, with teacher
supervision retained.

\subsection{Reward Design for Stepping Over Hazards}
\label{sec:step_over_rewards}

\begin{figure}[tbp]
    \centering
    \includegraphics[width=0.5\linewidth]{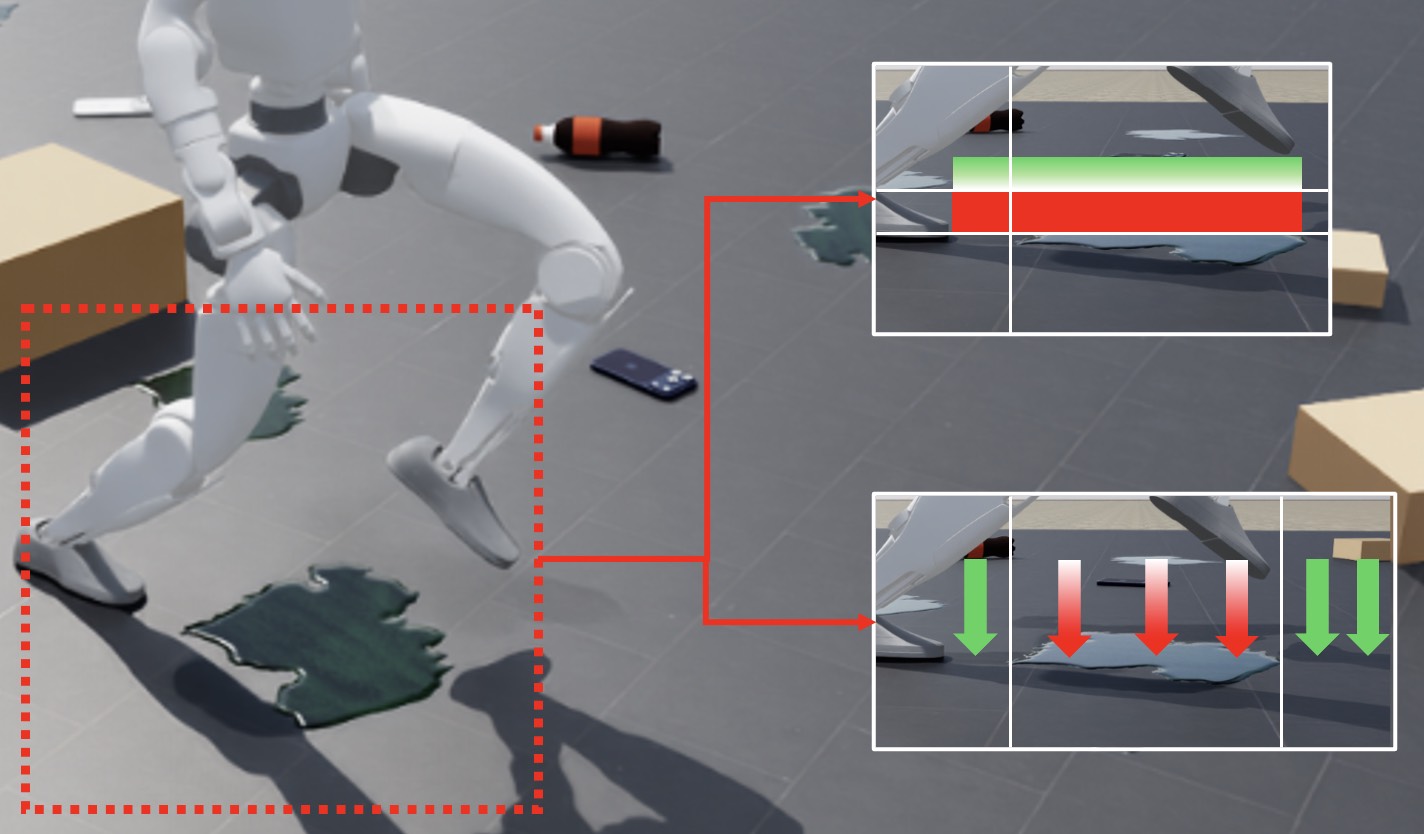}
    \caption{Two complementary reward terms for stepping over hazards. Top: rewarding sufficient foot clearance during horizontal swing motion. Bottom: penalizing premature descent while the swing foot still overlaps a hazard.}
    \label{fig:reward}
\end{figure}

We design two complementary reward terms for stepping over hazards: a
premature-descent penalty that discourages lowering the swing foot before
clearing a hazard, and a step-over clearance reward that encourages
sufficient foot clearance during horizontal swing motion.
Both terms are active only for a swing foot whose sole footprint overlaps
a lethal hazard, and are identically zero elsewhere, so the nominal gait
is untouched. For foot $i$, let $z_i$ be the lowest sole height and
$\mathbf{v}_i$ its velocity. Let $H_i$ and $h_i$ be the hazard overlap
indicator and the highest hazard top over the sole footprint---extended
by forward probes at $\{0,0.08,0.16\}\,\mathrm{m}$ in
\eqref{eq:step_over_reward}, current footprint only in
\eqref{eq:premature_descent_penalty}. Clearance is measured from the
contact-termination boundary, $d_i=z_i-(h_i+\epsilon)$ with
$\epsilon=0.03\,\mathrm{m}$, so neither term pays at a height that still
ends the episode. The gate
$g_i=\mathbb{I}[\|\mathbf{c}\|>0.1]\mathbb{I}[i\text{ in swing}]$
requires a live motion command, and $[\cdot]_0^1$ clips to $[0,1]$.
Here, $N_f$ is the number of feet.

\begin{equation}
\begin{aligned}
    r_{\mathrm{over}} = \frac{1}{N_f}\sum_{i=1}^{N_f}
    & g_i H_i
    \tanh\!\left(k\|\mathbf{v}_{i,xy}\|\right)
    \left[\frac{d_i}{\delta}\right]_0^1, \\
    & \delta=0.02\,\mathrm{m},\quad k=4.
\end{aligned}
    \label{eq:step_over_reward}
\end{equation}

\begin{equation}
\begin{aligned}
    p_{\mathrm{desc}} = {}& \frac{1}{N_f}\sum_{i=1}^{N_f}
    g_i H_i\,\mathbb{I}[d_i>0]
    \left[1-\frac{d_i}{D}\right]_0^1 \\
    & {}\times \tanh\!\left(
    \frac{\max(-v_{i,z},0)}{v_s}\right), \\
    & D=0.20\,\mathrm{m},\quad v_s=1.0\,\mathrm{m/s}.
\end{aligned}
    \label{eq:premature_descent_penalty}
\end{equation}

The horizontal-speed gate in \eqref{eq:step_over_reward} prevents a
stationary raised foot from earning clearance;
\eqref{eq:premature_descent_penalty} grows as a descending foot approaches
the boundary, below which termination applies instead.

\subsection{Anti-deviation for waypoint tracking in navigation}
\label{sec:anti_deviation}

\begin{figure}[tbp]
    \centering
    \includegraphics[width=0.45\linewidth]{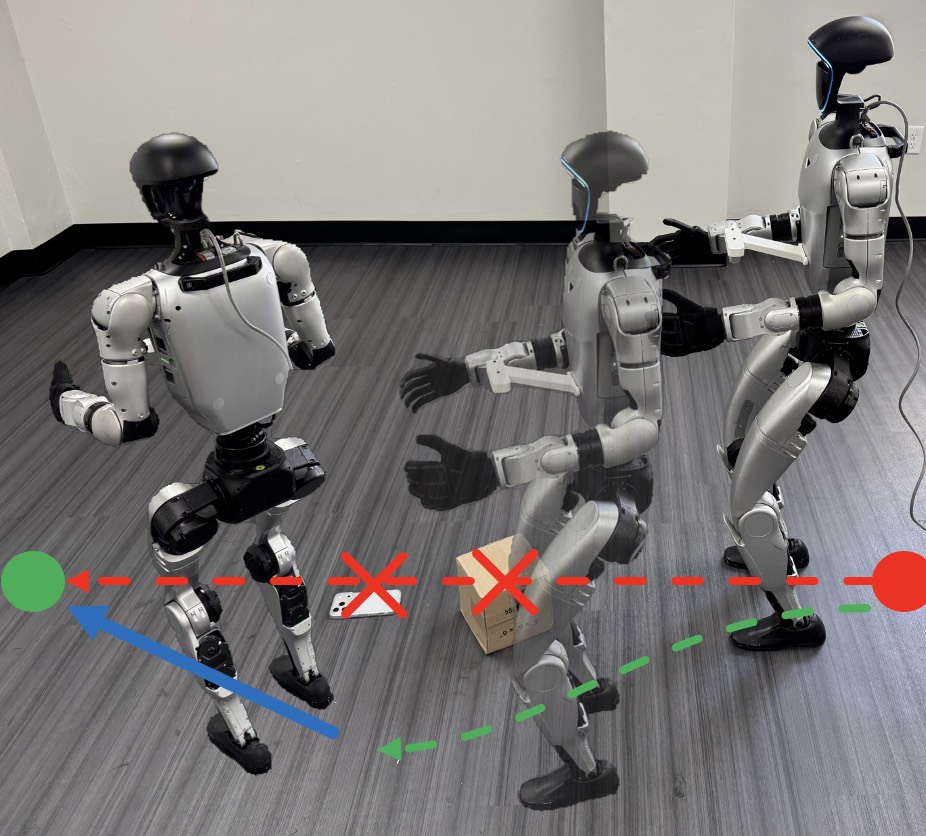}
    \caption{Command-execution mismatch. When no safe foothold is available along the direction toward the VLN-commanded waypoint (red), the robot detours around hazards, producing a deviated trajectory (green). The anti-deviation module detects this mismatch through localization feedback and adjusts the velocity commands to steer the robot back (blue) toward the intended target pose.}
    \label{fig:command_execution_gap}
\end{figure}

Figure~\ref{fig:command_execution_gap} illustrates how safety-aware locomotion
can cause the realized trajectory to deviate from the motion intended by
the high-level waypoint.

Without anti-deviation, $w_t$ is executed open-loop: it is converted into
a constant velocity command $c_t=w_t/\Delta t$ held for a fixed interval
$\Delta t$, and $\pi_{\mathrm{loco}}$ is left to track it. Deviations
introduced by hazard avoidance are then carried into the next VLN
decision. This open-loop execution is the ablation in
Table~\ref{tab:navigation_anti_deviation_ablation}.

To correct the mismatch between the target pose specified by the VLN
waypoint $w_t$ and the pose realized by $\pi_{\mathrm{loco}}$,
we introduce an anti-deviation module with localization feedback.
In simulation, LingBot-Map~\cite{chen2026geometric} estimates the current planar pose
$\hat p_t=(x_t,y_t,\theta_t)$ from the same monocular RGB stream
$o_t^{\mathrm{nav}}$ used by $\pi_{\mathrm{vln}}$.
The target planar pose $p_t^{\mathrm{tar}}=p_t\oplus w_t$ is anchored to
the robot's pose at the decision time, rather than recomputed from its
subsequent poses. The module compares $\hat p_t$ with this target pose
and maps the residual position and heading error to a compensatory
velocity command:
\begin{equation}
\begin{aligned}
    d_t &= p_t^{\mathrm{tar}}-\hat p_t,\\
    c_t^{\mathrm{comp}}
        &= \kappa_{\mathrm{ad}}(d_t)
         = \left(v_{x,t}^{\mathrm{comp}},v_{y,t}^{\mathrm{comp}},
         \psi_t^{\mathrm{comp}}\right).
\end{aligned}
    \label{eq:apd_command}
\end{equation}
Here, $\kappa_{\mathrm{ad}}$ denotes the anti-deviation control law that maps
the position and heading error $d_t$ to a compensatory velocity command.
The components $v_{x,t}^{\mathrm{comp}}$ and $v_{y,t}^{\mathrm{comp}}$ are the compensated
planar linear velocities, and $\psi_t^{\mathrm{comp}}$ is the compensated yaw rate.
The compensatory command replaces $c_t$ in the velocity-command slot of $r$,
while $\pi_{\mathrm{loco}}$ retains the same RGB-D inputs and action
interface. This feedback
steers the robot toward the intended target pose after local hazard
avoidance, limiting accumulated navigation drift.

\section{Experiments and Results}
\label{sec:experiments}
\subsection{Experiment Setup}
\label{sec:experiment_setup}
We evaluate humanoid locomotion in both simulation and physical-robot
experiments, while VLN is evaluated only in simulation with physically
simulated humanoid execution. Fig.~\ref{fig:testing_environments} provides
an overview of the three environment settings.

\noindent\textbf{Baselines.} Elevation-only is a student that retains only the
elevation branch of the AME architecture~\cite{he2025attention} and has no
semantic branch. It is distilled from the same robust semantic teacher
(Sec.~\ref{sec:semantic_branch}) with the same DAgger schedule
(Sec.~\ref{sec:rgbd_distillation}), the same depth-camera and proprioceptive
inputs, and the same perturbations as Ours; the only difference is that Ours
additionally receives RGB and DINOv2 features through the semantic branch.
This comparison isolates the contribution of semantic perception under identical
supervision.
The same Elevation-only checkpoint is used in simulation
and on the physical robot. Oracle is the semantic teacher with ground-truth
elevation and semantic maps. It is evaluated only under plant-side
perturbations (dynamics, payload, external disturbances, and proprioceptive
noise), bypassing perception perturbations: image noise, lighting, surface
colors, camera pose, and camera timing (frame hold and latency).

\subsubsection{Locomotion}
\emph{Simulation experiments.}
We evaluate the locomotion policy in procedurally generated corridor episodes. The
terrain suite comprises four groups: \emph{uneven ground},
\emph{stairs}, \emph{stakes}, and \emph{stonebridge} as summarized in
Table~\ref{tab:loco_terrain_groups}.

\begin{table}[!tb]
    \compacttable
    \setlength{\tabcolsep}{2pt}
    \renewcommand{\arraystretch}{0.95}
    \centering
    \caption{Terrain Groups and Difficulty Levels}
    \label{tab:loco_terrain_groups}
    \begin{tabular}{@{}p{0.21\columnwidth}p{\dimexpr0.79\columnwidth-4pt\relax}@{}}
        \toprule
        Group & Terrain; increasing difficulty (L1 $\rightarrow$ L3) \\
        \midrule
        Uneven Ground &
        Ground with obstacles, hazards, pits, and bumps;
        more/larger obstacles, greater height variation. \\
        Stairs &
        Descent/ascent from elevated/recessed starts; taller steps. \\
        Stakes &
        Isolated stake-top footholds over deep gaps;
        smaller footholds, wider gaps. \\
        Stonebridge &
        Narrow stepping stones without lateral support;
        narrower/shorter slabs, wider gaps. \\
        \bottomrule
    \end{tabular}
\end{table}

In addition to terrain and hazard variation, the locomotion benchmark includes
perturbations to dynamics, payload, external disturbances, sensing, visual
appearance, and camera mounting and timing. Table~\ref{tab:loco_perturbations}
summarizes their magnitudes at a perturbation scale of $1.0$.

\begin{table}[!tb]
    \compacttable
    \setlength{\tabcolsep}{2pt}
    \renewcommand{\arraystretch}{0.95}
    \centering
    \caption{Locomotion Perturbations at Scale $1.0$}
    \label{tab:loco_perturbations}
    \begin{tabular}{@{}p{0.21\columnwidth}p{\dimexpr0.79\columnwidth-4pt\relax}@{}}
        \toprule
        Category & Magnitudes at scale $1.0$ \\
        \midrule
        Dynamics &
        Friction: $0.4$--$1.0$; $k_p/k_d$: $\times 0.85$--$1.15$;
        motor: $\times 0.9$--$1.1$ \\
        Payload &
        Mass offset: $\pm 2$ kg; center-of-mass offset: $\pm 5$ cm \\
        Disturbances &
        Push: $\pm 0.8$ m/s every $3$--$7$ s;
        sustained force/torque: $\pm 5$ N / $\pm 1.67$ N$\cdot$m \\
        Proprio. noise &
        Angular vel.: $\pm 0.5$ rad/s; gravity proj.: $\pm 0.05$;
        joint pos./vel.: $\pm 0.05$ rad / $\pm 2$ rad/s \\
        Image noise &
        Depth: $\sigma = 2$ cm, $2\%$ dropout;
        RGB gain: $0.65$--$1.35$; RGB noise: $\sigma = 0.02$ \\
        Lighting &
        Dome/sun intensity: $450$--$1250$ / $0$--$1200$;
        color temp.: $3000$--$8000$ K \\
        Surface colors &
        Ground/deck luminance: $10$--$23$ / $10$--$60$ \\
        Camera pose &
        Translation: $\pm 5$ mm; rotation: $\pm 4^{\circ}$ \\
        Camera timing &
        Hold: $1$--$2$ control steps; latency: $60$--$100$ ms \\
        \bottomrule
    \end{tabular}
\end{table}

\emph{Physical experiments.} We evaluate locomotion on the Unitree G1 humanoid
robot equipped with the camera holder shown in Fig.~\ref{fig:methodology_overview}. We compare Elevation-only, Sonic~\cite{luo2025sonic} with human expert, and Ours at a fixed terrain configuration and
difficulty on plastic flooring, wooden boards, and a waterproof plastic sheet.
Sonic combines blind whole-body locomotion with velocity commands provided
by a human expert, who does not specify tracking points.
\subsubsection{Shared Hazard Types}
Table~\ref{tab:hazard_classes} lists the six hazard types shared by locomotion
and navigation.

\begin{table}[!tb]
    \compacttable
    \centering
    \caption{Hazard Types and Elevation Visibility}
    \label{tab:hazard_classes}
    \setlength{\tabcolsep}{1.5pt}
    \renewcommand{\arraystretch}{0.9}
    \resizebox{\columnwidth}{!}{%
    \begin{tabular}{@{}llcc@{}}
        \toprule
        Elevation Visibility & Hazard & Height (m) & Footprint (m) \\
        \midrule
        \multirow{2}{*}{Visible}
        & Crate & $0.06$--$0.24$ & $0.25$--$0.50\times0.30$--$0.60$ \\
        & Box & $0.05$--$0.18$ & $0.15$--$0.35\times0.15$--$0.35$ \\
        Ambiguous & Fallen bottle & $0.071$ & $0.24\times0.071$ \\
        \multirow{2}{*}{Invisible}
        & Phone & $0.014$ & $0.163\times0.079$ \\
        & Water/oil stain & $0$ & $0.10$--$0.45\times0.09$--$0.40$ \\
        \bottomrule
    \end{tabular}%
    }
\end{table}

\subsubsection{Navigation}
We evaluate navigation only in simulation, using the training and unseen
validation splits provided by
NaVILA VLN-CE-Isaac~\cite{cheng2025navila} and automatically augment the scenes
with script-generated obstacle placements, including hazards that are
geometrically visible, ambiguous or invisible. For each episode, we place $40\%$ of
the added obstacles on the desired navigation path and the remaining
$60\%$ within $1.0$~m of the path. By concentrating obstacles on and near the
reference route, this placement strategy creates a challenging benchmark
for hazard avoidance along both the intended route and nearby detours.

\subsection{Locomotion Result}
\subsubsection{Simulation Results}
We compare our semantic-aware policy (Ours) with Elevation-only and Oracle,
as defined in Sec.~\ref{sec:experiment_setup}.

\begin{table*}[!tb]
    \compacttable
    \scriptsize
    \setlength{\tabcolsep}{1.5pt}
    \centering
    \caption{Simulated Locomotion Across Difficulty and Perturbation Levels}
    \label{tab:loco_success}
    \resizebox{\textwidth}{!}{%
    \begin{tabular}{@{}ll*{12}{c}@{}}
        \toprule
        Method & Diff. & \multicolumn{3}{c}{Clean} & \multicolumn{3}{c}{Perturbation $0.5\times$} & \multicolumn{3}{c}{Perturbation $1.0\times$} & \multicolumn{3}{c}{Perturbation $1.5\times$} \\
        \cmidrule(lr){3-5}\cmidrule(lr){6-8}\cmidrule(lr){9-11}\cmidrule(lr){12-14}
        & & \shortstack{Success rate\\(SR) $\uparrow$} & Progress $\uparrow$ & \shortstack{Velocity tracking\\error (VTE) $\downarrow$} & \shortstack{Success rate\\(SR) $\uparrow$} & Progress $\uparrow$ & \shortstack{Velocity tracking\\error (VTE) $\downarrow$} & \shortstack{Success rate\\(SR) $\uparrow$} & Progress $\uparrow$ & \shortstack{Velocity tracking\\error (VTE) $\downarrow$} & \shortstack{Success rate\\(SR) $\uparrow$} & Progress $\uparrow$ & \shortstack{Velocity tracking\\error (VTE) $\downarrow$} \\
        \midrule
        \multirow{3}{*}{Oracle} & L1 & $99.0\pm1.3$ & $99.7\pm0.4$ & $0.103\pm0.001$ & $100.0\pm0.0$ & $100.0\pm0.0$ & $0.105\pm0.001$ & $98.8\pm0.8$ & $99.5\pm0.3$ & $0.116\pm0.002$ & $91.0\pm3.1$ & $97.3\pm1.4$ & $0.144\pm0.004$ \\
         & L2 & $95.6\pm4.0$ & $98.2\pm1.6$ & $0.108\pm0.002$ & $96.5\pm1.1$ & $98.1\pm0.9$ & $0.111\pm0.001$ & $93.3\pm0.8$ & $97.3\pm0.8$ & $0.127\pm0.003$ & $86.2\pm2.6$ & $94.6\pm2.0$ & $0.149\pm0.005$ \\
         & L3 & $95.0\pm0.8$ & $98.1\pm0.6$ & $0.113\pm0.001$ & $91.2\pm1.1$ & $96.7\pm0.5$ & $0.117\pm0.002$ & $86.2\pm2.5$ & $95.0\pm1.6$ & $0.132\pm0.002$ & $79.8\pm0.5$ & $93.0\pm0.8$ & $0.156\pm0.003$ \\
        \midrule
        \multirow{3}{*}{Elevation-only} & L1 & $47.9\pm4.0$ & $77.7\pm2.3$ & $0.137\pm0.004$ & $49.2\pm2.9$ & $77.4\pm2.7$ & $0.144\pm0.003$ & $40.6\pm3.4$ & $72.4\pm2.8$ & $0.167\pm0.004$ & $31.5\pm4.3$ & $65.2\pm3.1$ & $0.203\pm0.005$ \\
         & L2 & $32.7\pm5.3$ & $68.7\pm3.5$ & $0.155\pm0.006$ & $27.5\pm3.4$ & $66.5\pm3.1$ & $0.159\pm0.003$ & $20.4\pm4.1$ & $59.6\pm2.2$ & $0.182\pm0.004$ & $17.1\pm3.6$ & $54.9\pm2.6$ & $0.219\pm0.006$ \\
         & L3 & $25.2\pm4.7$ & $62.8\pm3.2$ & $0.168\pm0.002$ & $18.8\pm3.2$ & $60.2\pm2.3$ & $0.174\pm0.005$ & $16.0\pm2.7$ & $54.8\pm1.2$ & $0.199\pm0.004$ & $9.0\pm3.4$ & $49.7\pm1.6$ & $0.231\pm0.008$ \\
        \midrule
        \multirow{3}{*}{Ours} & L1 & $\mathbf{71.5\pm0.5}$ & $\mathbf{87.0\pm0.6}$ & $0.140\pm0.004$ & $\mathbf{70.4\pm2.8}$ & $\mathbf{85.9\pm1.9}$ & $0.145\pm0.003$ & $\mathbf{61.2\pm4.1}$ & $\mathbf{80.2\pm2.4}$ & $0.166\pm0.002$ & $\mathbf{46.9\pm5.0}$ & $\mathbf{72.0\pm2.2}$ & $0.201\pm0.005$ \\
         & L2 & $\mathbf{61.2\pm2.6}$ & $\mathbf{79.4\pm2.1}$ & $0.145\pm0.002$ & $\mathbf{54.2\pm5.0}$ & $\mathbf{76.5\pm3.2}$ & $0.157\pm0.002$ & $\mathbf{44.6\pm4.0}$ & $\mathbf{70.0\pm3.1}$ & $0.177\pm0.002$ & $\mathbf{27.1\pm3.7}$ & $\mathbf{60.5\pm1.7}$ & $0.214\pm0.005$ \\
         & L3 & $\mathbf{51.0\pm5.1}$ & $\mathbf{74.7\pm3.9}$ & $0.157\pm0.005$ & $\mathbf{47.1\pm8.1}$ & $\mathbf{72.1\pm4.7}$ & $0.166\pm0.003$ & $\mathbf{33.3\pm3.9}$ & $\mathbf{64.1\pm2.6}$ & $0.185\pm0.004$ & $\mathbf{22.7\pm1.8}$ & $\mathbf{56.7\pm1.9}$ & $0.220\pm0.002$ \\
        \bottomrule
    \end{tabular}%
    }
\end{table*}

Table~\ref{tab:loco_success} reports three complementary metrics. Success
rate (SR $\uparrow$) is the percentage of episodes that complete the full
terrain-specific target distance without leaving the corridor, falling, or
stepping on any obstacle or hazard. Foot contact with any hazard class---
elevation-visible, elevation-ambiguous, or elevation-invisible---counts as
failure, even if the robot remains upright. Progress ($\uparrow$) is the
farthest distance reached divided by the target distance, capped at one and
averaged over all episodes, expressed as a percentage; it captures partial
completion even in failed trials. Velocity tracking error (VTE $\downarrow$)
measures the discrepancy between commanded and realized velocity, capturing
command-following accuracy rather than task completion or safety.
Episodes terminate on hazard contact, fall, or leaving the corridor, so SR,
HCR, and FR sum to $100\%$ up to out-of-corridor terminations; these are
rare (at most $1.7\%$, Oracle at scale $1.5\times$) and are not reported
separately.

Ours achieves higher SR and Progress than Elevation-only across all difficulty
levels and perturbation settings. Under clean conditions, SR improves by
23.6, 28.5, and 25.8 percentage points on L1, L2, and L3, respectively,
while Progress increases by 9.3--11.9 percentage points. VTE remains broadly
comparable to Elevation-only, with slightly higher error in some settings.
Hazard avoidance can require sidesteps or detours away from the commanded
motion, increasing VTE even when these deviations improve safety.
This safety--tracking trade-off motivates our anti-deviation module,
which uses localization feedback to compensate for the resulting position
and heading errors during navigation, allowing local avoidance without
letting execution deviations accumulate into navigation drift.
(Table~\ref{tab:navigation_anti_deviation_ablation}).

Nevertheless, stronger perturbations and harder terrain reduce both SR and
Progress for Ours. On L3 at scale $1.5$, Ours achieves $56.7\%$ Progress
but only $22.7\%$ SR, highlighting the distinction between partial traversal
and completing the entire route without any failure.

\begin{table*}[!tb]
    \compacttable
    \setlength{\tabcolsep}{2pt}
    \setlength{\medmuskip}{1mu}
    \renewcommand{\arraystretch}{0.95}
    \centering
    \caption{Success and Failure Rates on Uneven Ground}
    \label{tab:loco_uneven_ground_failures}
    \resizebox{\textwidth}{!}{%
    \begin{tabular}{@{}lccccc@{\hspace{12pt}}lccccc@{}}
        \toprule
        Method & \multirow{2}{*}{\shortstack{Success rate\\(SR) $\uparrow$}} & \multicolumn{3}{c}{Hazard contact rate (HCR) $\downarrow$} & \multirow{2}{*}{\shortstack{Fall rate\\(FR) $\downarrow$}} & Method & \multirow{2}{*}{\shortstack{Success rate\\(SR) $\uparrow$}} & \multicolumn{3}{c}{Hazard contact rate (HCR) $\downarrow$} & \multirow{2}{*}{\shortstack{Fall rate\\(FR) $\downarrow$}} \\
        \cmidrule(lr){3-5}\cmidrule(lr){9-11}
        & & Visible & Ambiguous & Invisible & & & & Visible & Ambiguous & Invisible & \\
        \midrule
        \multicolumn{6}{c}{Clean} & \multicolumn{6}{c}{Perturbation $0.5\times$} \\
        Oracle & $93.6\pm3.0$ & $2.2\pm3.1$ & $1.7\pm2.7$ & $1.9\pm2.1$ & $0.6\pm0.7$ & Oracle & $92.2\pm3.4$ & $2.5\pm3.1$ & $0.8\pm0.7$ & $4.4\pm2.0$ & $0.0\pm0.0$ \\
        Elevation-only & $9.4\pm2.4$ & $27.8\pm7.7$ & $41.4\pm13.2$ & $21.1\pm11.6$ & $0.3\pm0.6$ & Elevation-only & $12.5\pm2.9$ & $29.4\pm9.6$ & $37.2\pm7.6$ & $20.3\pm5.5$ & $0.6\pm0.7$ \\
        Ours & $\mathbf{50.6\pm5.6}$ & $24.7\pm6.5$ & $17.2\pm5.8$ & $6.1\pm4.7$ & $1.4\pm1.2$ & Ours & $\mathbf{52.8\pm6.0}$ & $23.1\pm3.4$ & $16.4\pm5.9$ & $7.2\pm1.8$ & $0.6\pm0.7$ \\
        \midrule
        \multicolumn{6}{c}{Perturbation $1.0\times$} & \multicolumn{6}{c}{Perturbation $1.5\times$} \\
        Oracle & $86.9\pm3.0$ & $3.6\pm4.0$ & $2.2\pm1.1$ & $5.8\pm2.2$ & $0.8\pm1.1$ & Oracle & $72.8\pm6.1$ & $8.6\pm4.8$ & $4.4\pm2.4$ & $6.7\pm3.7$ & $5.8\pm2.7$ \\
        Elevation-only & $9.7\pm2.8$ & $33.3\pm7.6$ & $38.9\pm8.0$ & $17.5\pm4.7$ & $0.6\pm0.7$ & Elevation-only & $6.7\pm2.2$ & $38.6\pm9.2$ & $36.4\pm9.4$ & $15.0\pm3.4$ & $3.1\pm1.4$ \\
        Ours & $\mathbf{44.2\pm4.4}$ & $26.1\pm2.7$ & $16.7\pm4.1$ & $8.9\pm3.1$ & $3.6\pm2.3$ & Ours & $\mathbf{28.1\pm5.7}$ & $31.9\pm5.6$ & $20.0\pm5.9$ & $6.7\pm1.6$ & $11.9\pm2.7$ \\
        \bottomrule
    \end{tabular}%
    }
\end{table*}

Table~\ref{tab:loco_uneven_ground_failures} evaluates locomotion on uneven ground, with a primary focus on obstacle and hazard avoidance. Ours outperforms
Elevation-only in SR under all four conditions, with gains of 41.2, 40.3, 34.5, and
21.4 percentage points under clean conditions and perturbation scales of
$0.5$, $1.0$, and $1.5$, respectively.

To quantify how much of the Elevation-only--Oracle gap is closed by Ours,
we compute $(\mathrm{HCR}_{\mathrm{Elevation\text{-}only}}-
\mathrm{HCR}_{\mathrm{Ours}})/(\mathrm{HCR}_{\mathrm{Elevation\text{-}only}}-
\mathrm{HCR}_{\mathrm{Oracle}})$ for each hazard class and condition in
Table~\ref{tab:loco_uneven_ground_failures}. Using the reported mean HCRs, Ours closes
approximately $74$--$100\%$ of the gap for elevation-invisible hazards,
$51$--$61\%$ for elevation-ambiguous hazards, and only $12$--$24\%$ for
elevation-visible hazards. The semantic branch therefore recovers a larger
fraction of the gap where elevation cues are weak or absent, while much of
the gap on geometrically visible hazards remains. These ratios describe
class-specific HCR improvements, not recovery of overall success rate.

\subsubsection{Physical results}
Table~\ref{tab:loco_physical} reports Clean SR ($\uparrow$), the
out-of-corridor rate ($\downarrow$), and hazard contact rates ($\downarrow$)
for elevation-map-visible, -ambiguous, and -invisible obstacles, all in $\%$.
We conduct five trials for each method on each flooring type.
Clean SR is the percentage of trials completed without falls, departures
from the corridor, or contact with obstacles or hazards.
Contacts with obstacles or hazards are counted for the contact metrics but
do not terminate a trial; the robot continues executing the task, although
the trial no longer qualifies as a clean success. Our policy
incurs no falls in these trials.

\begin{table}[!htbp]
    \compacttable
    \setlength{\tabcolsep}{1pt}
    \centering
    \caption{Real-World Locomotion on Three Surfaces}
    \label{tab:loco_physical}
    \begin{tabular*}{\columnwidth}{@{\extracolsep{\fill}}l*{5}{c}@{}}
        \toprule
        Method & \shortstack{Clean SR\\$\uparrow$} & \shortstack{Out of\\corridor $\downarrow$} & \multicolumn{3}{c}{Hazard Contact Rate(HCR) $\downarrow$} \\
        \cmidrule(lr){4-6}
        & & & Visible & Ambiguous & Invisible \\
        \midrule
        \multicolumn{6}{c}{\textbf{Plastic flooring} (dark-colored, very slippery)} \\
        Elevation-only & 0 & 20 & 20 & 80 & 100 \\
        Sonic & 20 & 80 & 0 & 40 & 40 \\
        Ours & 40 & 20 & 0 & 20 & 20 \\
        \midrule
        \multicolumn{6}{c}{\textbf{Wooden boards} (light brown, rough, narrow)} \\
        Elevation-only & 0 & 0 & 20 & 80 & 100 \\
        Sonic & 0 & 100 & 0 & 20 & 20 \\
        Ours & 60 & 40 & 0 & 0 & 20 \\
        \midrule
        \multicolumn{6}{c}{\textbf{Waterproof plastic sheet} (light gray, slippery, narrow)} \\
        Elevation-only & 0 & 20 & 0 & 20 & 100 \\
        Sonic & 0 & 100 & 0 & 20 & 20 \\
        Ours & 60 & 20 & 0 & 20 & 0 \\
        \bottomrule
    \end{tabular*}
\end{table}

Even with expert-provided velocity commands, Sonic, which lacks environmental perception, leaves the corridor
in $80$--$100\%$ of trials while avoiding hazards. In these confined
settings, steering through high-level velocity commands alone is insufficient for reliable avoidance
without hazard-aware foot placement.

\emph{Robustness to physical disturbances.}
Table~\ref{tab:loco_physical_disturbances} compares Ours with no disturbance
and under three physical disturbance conditions. Consistent with the
simulation results, our policy shows robustness to the tested common
disturbances.

\begin{table}[!htbp]
    \compacttable
    \centering
    \caption{Robustness to Physical Disturbances}
    \label{tab:loco_physical_disturbances}
    \begin{tabular*}{\columnwidth}{@{\extracolsep{\fill}}lc@{}}
        \toprule
        Condition & Clean SR $\uparrow$ \\
        \midrule
        No disturbance & 60 \\
        Flashlight (2000~lm) & 40 \\
        Added payload (1Kg) & 40 \\
        Pool-noodle strikes & 60 \\
        \bottomrule
    \end{tabular*}
\end{table}

\subsection{Navigation}

We run two controlled ablations, removing the semantic branch
(Elevation-only) or disabling anti-deviation while keeping all other
components unchanged. SR requires reaching the goal without collision or
fall but permits contact with water or oil stains. Safe SR additionally
requires no hazard intrusion (any foot entering a hazard region). Progress
measures normalized geodesic goal-distance reduction; SPL weights success
by path efficiency; FGD is the final geodesic distance to the goal.

\begin{table*}[!tb]
    \compacttable
    \centering
    \caption{Semantic-Branch Ablation in Simulated Navigation}
    \label{tab:navigation_simulation_ablation}
    \resizebox{\textwidth}{!}{%
    \begin{tabular}{@{}ll*{20}{c}@{}}
        \toprule
        Method & \shortstack{Perturbation\\scale} & \multicolumn{10}{c}{5 obstacles} & \multicolumn{10}{c}{10 obstacles} \\
        \cmidrule(lr){3-12}\cmidrule(lr){13-22}
        & & Safe SR $\uparrow$ & \shortstack{Avg.\\Progress $\uparrow$} & \shortstack{Collision\\Rate $\downarrow$} & \shortstack{Hazard\\Intrusion $\downarrow$} & FR $\downarrow$ & SR $\uparrow$ & SPL $\uparrow$ & \shortstack{Final goal\\distance (m) $\downarrow$} & \shortstack{VLM early\\stop $\downarrow$} & Timeout $\downarrow$ & Safe SR $\uparrow$ & \shortstack{Avg.\\Progress $\uparrow$} & \shortstack{Collision\\Rate $\downarrow$} & \shortstack{Hazard\\Intrusion $\downarrow$} & FR $\downarrow$ & SR $\uparrow$ & SPL $\uparrow$ & \shortstack{Final goal\\distance (m) $\downarrow$} & \shortstack{VLM early\\stop $\downarrow$} & Timeout $\downarrow$ \\
        \midrule
        Ours full & $0.5$ & $35.38\%$ & $82.46\%$ & $15.04\%$ & $11.98\%$ & $16.16\%$ & $40.95\%$ & $0.385$ & $1.67$ & $0.56\%$ & $27.30\%$ & $17.55\%$ & $43.34\%$ & $20.89\%$ & $18.11\%$ & $11.98\%$ & $24.23\%$ & $0.212$ & $5.58$ & $24.79\%$ & $18.11\%$ \\
        Elevation-only & $0.5$ & $23.12\%$ & $72.13\%$ & $22.84\%$ & $20.61\%$ & $25.07\%$ & $30.08\%$ & $0.250$ & $3.05$ & $3.62\%$ & $18.38\%$ & $13.65\%$ & $48.04\%$ & $29.25\%$ & $27.86\%$ & $21.73\%$ & $21.17\%$ & $0.155$ & $5.60$ & $16.71\%$ & $11.14\%$ \\
        \midrule
        Ours full & $1.0$ & $23.68\%$ & $43.20\%$ & $12.26\%$ & $12.81\%$ & $10.86\%$ & $29.80\%$ & $0.256$ & $5.38$ & $29.24\%$ & $17.83\%$ & $13.09\%$ & $49.78\%$ & $32.59\%$ & $22.28\%$ & $14.21\%$ & $20.06\%$ & $0.178$ & $4.55$ & $14.48\%$ & $18.66\%$ \\
        Elevation-only & $1.0$ & $17.27\%$ & $39.89\%$ & $20.89\%$ & $22.84\%$ & $19.50\%$ & $23.96\%$ & $0.185$ & $6.20$ & $25.07\%$ & $10.58\%$ & $9.75\%$ & $42.00\%$ & $33.43\%$ & $33.43\%$ & $23.68\%$ & $16.16\%$ & $0.125$ & $6.00$ & $13.37\%$ & $13.37\%$ \\
        \bottomrule
    \end{tabular}%
    }

\end{table*}

Table~\ref{tab:navigation_simulation_ablation} highlights three findings.
Across the four evaluated settings, semantic input raises Safe SR by
$3.34$--$12.26$ percentage points and reduces hazard intrusion and FR by
$8.63$--$11.15$ and $8.64$--$9.75$ points, respectively. Collision rate also
decreases in every setting, although the reduction narrows to $0.84$ points
with ten obstacles at scale $1.0$. Progress improves in three settings; the
exception is ten obstacles at scale $0.5$, where Ours trades $4.7$ points of
Progress for lower collision, hazard-intrusion, and fall rates. Increasing
the obstacle count generally lowers Safe SR, while the gaps between SR and
Safe SR show why goal completion alone is insufficient for evaluating safe
humanoid navigation.

\begin{table*}[!tb]
    \compacttable
    \centering
    \caption{Anti-Deviation Ablation in Simulated Navigation}
    \label{tab:navigation_anti_deviation_ablation}
    \resizebox{\textwidth}{!}{%
    \begin{tabular}{@{}l*{10}{c}@{}}
        \toprule
        Method & Safe SR $\uparrow$ & \shortstack{Avg.\\Progress $\uparrow$} & \shortstack{Collision\\Rate $\downarrow$} & \shortstack{Hazard\\Intrusion $\downarrow$} & FR $\downarrow$ & SR $\uparrow$ & SPL $\uparrow$ & \shortstack{Final goal\\distance (m) $\downarrow$} & \shortstack{VLM early\\stop $\downarrow$} & Timeout $\downarrow$ \\
        \midrule
        Ours & $23.68\%$ & $43.20\%$ & $12.26\%$ & $12.81\%$ & $10.86\%$ & $29.80\%$ & $0.256$ & $5.38$ & $29.24\%$ & $17.83\%$ \\
        Ours w/o anti-deviation & $11.98\%$ & $46.12\%$ & $35.93\%$ & $18.38\%$ & $8.64\%$ & $19.50\%$ & $0.169$ & $5.16$ & $18.38\%$ & $17.55\%$ \\
        \bottomrule
    \end{tabular}%
    }
\end{table*}

Table~\ref{tab:navigation_anti_deviation_ablation} shows that anti-deviation
substantially improves successful and safe completion. It raises Safe SR from
$11.98\%$ to $23.68\%$, SR from $19.50\%$ to $29.80\%$, and SPL from
$0.169$ to $0.256$. Collision rate falls from $35.93\%$ to $12.26\%$
and hazard intrusion from $18.38\%$ to $12.81\%$, while FR remains broadly
similar ($8.64\%$ versus $10.86\%$). Notably, these substantial gains in SR
and SPL occur despite slightly lower average Progress ($46.12\%$ to
$43.20\%$). Since Progress measures normalized geodesic goal-distance
reduction, an agent can make progress toward the goal without reliably
tracking the intended waypoints or completing the task. This contrast
suggests that, without continuous execution correction, the robot may
still move generally toward the destination while accumulating deviations
that prevent successful arrival. By using localization feedback to
continually correct position and heading errors relative to the intended
target pose, anti-deviation helps translate partial progress into successful,
path-efficient completion. Final goal distance remains similar ($5.16$
versus $5.38$~m), while VLM early stops increase ($18.38\%$ to $29.24\%$)
and timeout remains similar, indicating that the benefits are concentrated
in successful and safe completion rather than uniform improvements across
all metrics.

\subsection{Limitations}
Semantic hazard avoidance does not eliminate
the teacher--student performance gap. Even on uneven ground under clean conditions, Ours
achieves $50.6\%$ SR versus Oracle's $93.6\%$, a $43.0$-percentage-point
gap, and its elevation-visible HCR remains $24.7\%$ versus $2.2\%$.
Thus, improved handling of ambiguous and invisible hazards is not sufficient
for reliable route completion. Because Oracle and Ours differ in both their
observation interfaces and teacher--student training roles, these results
do not isolate perception error from distillation or control limitations.
Under perturbation scale $1.5$, Ours' FR rises to $11.9\%$ compared with
Oracle's $5.8\%$ and Elevation-only's $3.1\%$. This increase may reflect
the camera's sensitivity to changes in lighting conditions.

\section{Conclusion}

This work presents a benchmark and semantic-aware locomotion-navigation framework that address two gaps in humanoid navigation: hazards poorly represented in elevation maps and deviations between commanded and realized motion. On hazards that are ambiguous or invisible in elevation maps, a depth-only student distilled from the same teacher contacts them roughly $2$-$3\times$ more often than our semantic-aware student, which closes $51$-$61\%$ and $74$-$100\%$ of the hazard-contact-rate gap to the ground-truth-map oracle in these two classes, respectively, but only $12$-$24\%$ on geometrically visible hazards. These results suggest that semantic information addresses important geometric blind spots, although the residual gap cannot be attributed to perception coverage alone. On the physical Unitree G1, the same policy traverses cluttered floors with water spills, phones, and fallen bottles where the depth-only baseline fails. In navigation, anti-deviation raises completion from $19.5\%$ to $29.8\%$ SR while reducing hazard intrusion from $18.38\%$ to $12.81\%$ (Table~\ref{tab:navigation_anti_deviation_ablation}), demonstrating the benefit of closed-loop execution correction.

Future work will extend the benchmark to model the physical effects of hazards, such as slipping and unstable contact. Persistent semantic mapping could retain previously observed hazards outside the waist camera's current field of view. Another direction is to incorporate hazard awareness into target-pose selection for anti-deviation, balancing trajectory recovery with safe locomotion. We also plan to validate the complete navigation framework on hardware. These directions would strengthen both the system and its evaluation, advancing humanoid navigation research field that accounts jointly for task completion, hazard avoidance, and execution consistency.

\bibliographystyle{plain}
\bibliography{references}

\end{document}